# Expanding the Lexicon of Ge'ez Based African Languages: A Comparative Study of Amharic and Tigrinya

**Hailay Kidu Teklehaymanot**[†], **Debela Desalegn Yadeta**[‡], **Wolfgang Nejdl**[†]
[†] L3S Research Center, Leibniz University Hannover, Germany {teklehaymanot, nejdl}@l3s.de
[‡] Addis Ababa University, Ethiopia debela.desalegn@aau.edu.et

## Abstract

Multilingual pre-trained language models such as XLM-R perform well for major languages but struggle with low-resource Ge'ez-script languages, largely because Latin-script-centric tokenizers split their words into many subwords. We introduce VEXMLM, a vocabulary-extended variant of XLM-R targeting Amharic and Tigrinya. We train language-specific SentencePiece tokenizers on monolingual corpora, extend XLM-R's vocabulary with 30k Ge'ez-script subwords, and initialize each new embedding to the mean of the pretrained embeddings. VEXMLM undergoes two-stage training: (1) continued masked language modeling on the monolingual corpora and (2) supervised fine-tuning on question answering and named entity recognition (Amharic and Tigrinya) and sentiment analysis (Amharic). VEXMLM lowers tokenizer fertility below that of XLM-R and Glot500 on both languages, by 28.0% (Amharic) and 45.9% (Tigrinya) relative to XLM-R. Downstream, it modestly improves named entity recognition over XLM-R, scores below XLM-R on extractive question answering, and is comparable on sentiment analysis. An ablation on Tigrinya NER shows that vocabulary expansion alone lowers accuracy on out-of-vocabulary words (words that XLM-R's tokenizer cannot represent or splits into more pieces than the expanded tokenizer), and that continued pretraining is required for the expanded model to exceed the baseline. Vocabulary expansion thus makes Ge'ez-script tokenization substantially more efficient, while its downstream benefit depends on the task and on adapting the new embeddings through continued pretraining. **Resources:** GitHub repository | Hugging Face model

## 1 Introduction

Multilingual pre-trained language models (PLMs) have significantly advanced natural language processing (NLP) by enabling cross-lingual transfer across many languages (Wang et al., 2023; Xue et al., 2021). Despite these advances, low-resource languages remain severely underrepresented in training corpora, leading to suboptimal performance on downstream tasks. This underrepresentation is particularly acute for languages that use non-Latin scripts, where standard tokenization strategies allocate disproportionately few subword units, resulting in high out-of-vocabulary (OOV) rates and excessive token fragmentation that degrades performance on token-sensitive tasks such as named entity recognition, sequence labeling, and reading comprehension (Chau and Smith, 2021; Ahia et al., 2023).

A key technical bottleneck is subword tokenization. Standard methods such as Byte Pair Encoding (BPE) and SentencePiece (Kudo and Richardson, 2018) are highly effective for high-resource, Latin-script languages but frequently over-segment text in morphologically rich or orthographically distinct languages, producing longer sequences, degraded semantic representations, and elevated OOV rates. Tokens not covered by the model's vocabulary are typically mapped to a generic [UNK] symbol, discarding all lexical information (Liu et al., 2024; Xue et al., 2022).

These challenges are especially pronounced for Ge'ez-script languages such as Amharic and Tigrinya. The Ge'ez script is a morphosyllabic abugida in which each symbol represents a consonant vowel pair, which gives it a large character inventory (Getatchew Haile, 1996). Standard multilingual vocabularies built predominantly on Latin-script corpora inevitably produce severe fragmentation for these scripts. Compounding this problem, Amharic and Tigrinya have limited annotated digitized text, orthographic variation across dialects, and inconsistent Unicode encoding in existing datasets, all of which further hamper model adaptation.

Prior work has explored vocabulary expansion for low-resource languages (Wang et al., 2019), con-tinued pretraining on parallel corpora (Ebrahimi and Kann, 2021), adaptation strategies for African languages (Alabi et al., 2022), and large-scale pretraining over 500+ languages (ImaniGooghari et al., 2023). However, to our knowledge none of these approaches has been systematically applied to Ge'ez-script languages, and no prior work has combined targeted vocabulary augmentation, language-specific tokenizer construction, mean-based embedding initialization, and two-stage continued pretraining within a unified framework for this language family. Existing broad-coverage models such as Glot500 achieve coverage through scale but do not optimize tokenization for specific script families, leaving a representational gap for Amharic and Tigrinya.

This paper addresses these limitations by proposing **VEXMLM** (Vocabulary-Extended XLM-R), a systematic framework for adapting XLM-R to Ge'ez-script languages that combines: (1) language-specific SentencePiece tokenization and vocabulary augmentation with 30,000 Ge'ez-derived subword tokens; (2) mean initialization of new token embeddings to preserve alignment with the existing embedding space; and (3) a two-stage training procedure comprising continued masked language model pretraining followed by task-specific fine-tuning. The adapted model is evaluated on three downstream tasks: named entity recognition (NER) and question answering (QA) in Amharic and Tigrinya, and sentiment analysis (SA) in Amharic.

This work makes three principal contributions:

- **Vocabulary expansion and embedding initialization for Ge'ez script.** We propose a vocabulary augmentation approach, constructing language-specific SentencePiece tokenizers for Amharic and Tigrinya and initializing new token embeddings via the mean of source embeddings, ensuring compatibility with the pretrained XLM-R embedding space.

- **Two-stage training.** We show that continued MLM pretraining on Ge'ez-script monolingual corpora is required for the expanded vocabulary to help: without it, vocabulary expansion lowers Tigrinya NER accuracy on out-of-vocabulary words, and with it the model exceeds the XLM-R baseline.

- **Intrinsic and extrinsic evaluation.** We evaluate VEXMLM with tokenization efficiency metrics (fertility, compression, OOV round-trip) and with downstream performance on NER and QA in Amharic and Tigrinya and sentiment analysis in Amharic, reporting means and standard deviations over five seeds and including the tasks on which VEXMLM does not improve over XLM-R.

## 2 Background and Related Work

### 2.1 The Ge'ez Script

The Ge'ez script, known natively as *fidäl*, stands as the only indigenous writing system of the African continent to have evolved continuously into a modern, productive form. With its earliest attested inscriptions traced to the $4^{th}$ century CE, exemplified by the Hawulti obelisk in Matara, Eritrea, the script emerged within the epigraphic traditions of the Aksumite civilization of the northern Horn of Africa and represents a sophisticated adaptation of earlier South Semitic conventions (Phillipson, 2012; Getatchew Haile, 1996). Unlike the alphabetic systems of Greek or Latin, Ge'ez is fundamentally a syllabic abugida: each character encodes a consonant-vowel unit (Getatchew Haile, 1996). This structural innovation arose through deliberate transformation of the South Arabian Sabean script, which Ge'ez adopted and modified by introducing a vowel modification system and shifting the writing direction from right-to-left to left-to-right (Getatchew Haile, 1996; Gebremedhin and Mebrahtu, 2020). This morphological complexity and script-level unity across multiple languages (Amharic, Tigrinya, Tigre, and Classical Ge'ez) create a unique computational challenge.

Amharic adopted the base Ge'ez characters and added characters for additional phonemes, including Cushitic sounds encoded via modified Ge'ez forms. Tigrinya retained the full Ge'ez character set and added eight additional *fidels* (Gebremedhin and Mebrahtu, 2020). These script properties (large character inventories, morphosyllabic structure, and limited presence in multilingual corpora) directly motivate the need for language-specific tokenization.

### 2.2 Vocabulary Expansion in Multilingual Models

Vocabulary expansion methods aim to augment the tokenizer and embedding matrix of a pretrained multilingual model to better serve target languages underrepresented during pretraining. Wang et al.

(2019) identified fixed vocabulary size as a structural bottleneck in multilingual models and proposed vocabulary adaptation strategies. More recently, Kim et al. (2024) demonstrated vocabulary expansion for Korean (EEVE-Korean) via parameter freezing and subword embedding initialization, and reported strong downstream performance. Yamaguchi et al. (2024) examined the minimal data requirements for vocabulary expansion, and Hong et al. (2024) proposed language-specific model heads with customized vocabulary sets combined with verification-based fine-tuning.

Remy et al. (2024) introduced Trans-Tokenization, a framework for repurposing high-resource models for low-resource settings via adaptive tokenization. Chelombitko et al. (2024) presented QTok, a comprehensive framework for evaluating multilingual tokenization quality across languages.

This work differs from these efforts in two key respects. First, we focus on Ge'ez-script languages, which have received little systematic attention in the vocabulary expansion literature despite their significant representational challenges. Second, we combine vocabulary expansion with a two-stage pretraining and fine-tuning pipeline and isolate the contribution of each stage in an ablation.

### 2.3 Tokenization for Morphologically Rich Languages

Subword tokenizers (WordPiece, BPE, SentencePiece) reduce vocabulary size while maintaining generalization across common forms. However, in low-resource settings, particularly for morphologically rich languages, these tokenizers suffer from over-segmentation, high OOV rates, and data inefficiency (Limisiewicz et al., 2023; Beinborn and Pinter, 2023). The problem is amplified for languages with complex derivational or inflectional morphology (Chai et al., 2024).

Vocabulary-free approaches offer an alternative: ByT5 (Xue et al., 2022) and CANINE (Clark et al., 2022) operate at the byte or character level and demonstrate robustness across many languages. However, these models incur higher computational costs due to longer sequences and slower inference (Xue et al., 2021; Liang et al., 2023). Pagnoni et al. (2024) proposed the Byte Latent Transformer (BLT), which processes raw bytes without tokenization, representing a promising but computationally demanding alternative.

### 2.4 Multilingual Models for Languages in Africa

Most multilingual modeling research has concentrated on high-resource languages. Some efforts have expanded coverage: ImaniGooghari et al. (2023) proposed Glot500-m, which covers 511 predominantly low-resource languages, significantly expanding the set of supported languages. Alabi et al. (2022) fine-tuned XLM-R for languages in Africa. AfroXLMR (Alabi et al., 2022) and AfriBERTa (Ogueji et al., 2021) provide specialized encoders for some languages in Africa.

Despite this progress, Ge'ez-script languages such as Amharic and Tigrinya remain underserved in terms of tokenization optimization and vocabulary coverage. This work directly targets this gap by constructing language-specific tokenizers, augmenting XLM-R's vocabulary with Ge'ez-script (Amharic and Tigrinya) subword tokens, and evaluating the resulting model on Amharic and Tigrinya downstream tasks.

## 3 Proposed Method

This section describes the VEXMLM framework for mitigating over-segmentation of Ge'ez-script languages within a multilingual PLM. The framework consists of four components: (1) base model selection, (2) language-specific tokenizer construction and vocabulary expansion/integration, (3) mean-based embedding initialization for new tokens, and (4) a two-stage training procedure of continued pretraining followed by task-specific fine-tuning.

### 3.1 Multilingual Base Model

We adopt XLM-RoBERTa (XLM-R) (Conneau et al., 2020) as the base model due to its strong cross-lingual transfer capabilities. XLM-R is built on the Transformer architecture (Vaswani et al., 2017) and the RoBERTa framework (Liu et al., 2019), an optimized variant of BERT (Devlin et al., 2019), and is pretrained via masked language modeling (MLM) on CommonCrawl data spanning 100 languages.

Despite broad coverage, XLM-R exhibits systematic limitations for Ge'ez-script languages. Its Latin-centric SentencePiece vocabulary contains few Ge'ez-script subwords, so Amharic and Tigrinya words are split into many pieces (Zhang et al., 2022; Ahia et al., 2023); on our data, XLM-R produces 2.0692 (Amharic) and 3.1300 (Tigrinya) tokens per word, although it can still represent nearly every word without an [UNK] token (Table 2). This over-segmentation motivates our targeted augmentation strategy.

### 3.2 Target Language Tokenization and Vocabulary Expansion

We construct independent language-specific SentencePiece unigram tokenizers (Kudo and Richardson, 2018) for Amharic and Tigrinya, trained on the training portions of the monolingual corpora described in Section 4.3 (126,814 Amharic and 137,772 Tigrinya lines; NFC normalization, character coverage 0.9995). For Tigrinya, we train a vocabulary of 50,000 subword units; for Amharic, 32,000 subword units.

To merge the new vocabulary with XLM-R's original vocabulary $\mathcal{V}^s$, we consider only pieces of the new tokenizers that are absent from $\mathcal{V}^s$ (27,698 Amharic and 47,072 Tigrinya pieces). From each language we take the 20,000 most frequent of these pieces in its training corpus, interleave the two ranked lists, and remove cross-language duplicates until 30,000 pieces are selected: 14,189 Amharic-only, 14,279 Tigrinya-only and 1,532 shared pieces. The selected pieces are merged into XLM-R's SentencePiece model rather than appended as separate added tokens, so that they are segmented by the same unigram model as the original vocabulary. This expands XLM-R's vocabulary from 250,002 to 280,002 entries. Backward compatibility with all original XLM-R representations is fully preserved, as no existing entries are modified.

### 3.3 Embedding Initialization for Expanded Vocabulary

Integrating new vocabulary tokens into a pretrained model requires initializing their embeddings compatibly with the existing embedding space. Random initialization risks destabilizing pretraining dynamics, while zero-initialization introduces systematic magnitude bias. We adopt a simple *mean initialization* strategy, which positions every new token at the centroid of the source embedding distribution. More informed methods such as WECHSEL (Minixhofer et al., 2022) instead initialize each new embedding as a similarity-weighted combination of its nearest source embeddings; we do not compare against them.

**Formal Setup.** Let $\theta$ denote the parameters of the pretrained model $LM^s_\theta$, trained on source vocabulary $\mathcal{V}^s = \{v^s_1, \ldots, v^s_n\}$ with embeddings $\mathcal{E}^s = \{e^s_1, \ldots, e^s_n\}$, where $e^s_i \in \mathbb{R}^d$. The probability of predicting token $w_i$ given context $w_{1:i-1}$ is:

$$p_\theta(w_i \mid w_{1:i-1}) = \frac{\exp(h_{i-1}^\top e^s_{w_i})}{\sum_{j \in \mathcal{V}^s} \exp(h_{i-1}^\top e^s_j)}, \quad (1)$$

where $h_{i-1} = \phi(w_{1:i-1}; LM^s_\theta) \in \mathbb{R}^d$ is the encoder's contextual representation.

**Vocabulary Extension.** Let $\mathcal{V}^t = \{v^t_1, \ldots, v^t_{n'}\}$ be the set of new tokens with $\mathcal{V}^t \cap \mathcal{V}^s = \emptyset$ by construction.[1] The updated parameter set becomes $\theta' = \theta \cup \{e^t_j : j \in \mathcal{V}^t\}$, and the extended model $LM^t_{\theta'}$ operates over $\mathcal{V}^{s \cup t}$.

**Mean Initialization.** Each new embedding $e^t_j$ is initialized to the mean of all source embeddings:

$$e^t_j = \frac{1}{|\mathcal{V}^s|} \sum_{i=1}^{|\mathcal{V}^s|} e^s_i. \quad (2)$$

This places new tokens at the centroid of the source embedding space, reducing the risk of outlier placement and avoiding directional bias toward any specific source language. All new tokens therefore start from the same vector and are differentiated only during continued pretraining. Since XLM-R ties its input embedding matrix with the MLM output projection, Equation 2 is applied once to the shared matrix. An ablation comparing mean initialization against random initialization is reported in Section 5.3.

### 3.4 Two-Stage Training: Continued Pretraining and Fine-Tuning

**Stage 1: Continued Pretraining.** We perform continued pretraining using the MLM objective ($p_{\text{mask}} = 0.15$) on the Amharic and Tigrinya monolingual corpora. During this stage, *all* model parameters, including original and newly added embeddings, are updated. The two languages are resampled with probabilities proportional to $n_\ell^\alpha$, where $n_\ell$ is the number of training lines of language $\ell$ and $\alpha = 0.50$, which yields 129,552 Amharic and 135,034 Tigrinya lines. Full parameter training allows newly initialized embeddings to move toward meaningful representations within the shared embedding space. The model at the end of this stage is referred to as **VEXMLM**.

[1] Any token already present in $\mathcal{V}^s$ retains its original embedding; only genuinely new tokens enter $\mathcal{V}^t$.

| Hyperparameter | Cont. Pretraining | Fine-Tuning |
|---|---|---|
| Maximum sequence length | 256[†] | 256 |
| Batch size | 32 | 32 |
| Training epochs | 60 (best: 56) | 4 |
| Learning rate | $5 \times 10^{-5}$ | $2 \times 10^{-5}$ |
| LR schedule | Linear, 6% warmup | Linear, 10% warmup |
| Weight decay | 0.01 | 0.01 |
| MLM probability | 0.15 | N/A |
| Gradient clipping | 1.0 | 1.0 |
| Optimizer | AdamW | AdamW |
| Adam $\epsilon$ | $1 \times 10^{-8}$ | $1 \times 10^{-8}$ |
| Adam $\beta_1$ | 0.9 | 0.9 |
| Adam $\beta_2$ | 0.999 | 0.999 |
| Mixed precision | bf16 | bf16 |
| Trainable parameters | All | All |

Table 1: Hyperparameter settings for continued pretraining and supervised fine-tuning. For pretraining, the checkpoint with the lowest validation loss (epoch 56 of 60) is released.

**Stage 2: Task-Specific Fine-Tuning.** VEXMLM is subsequently fine-tuned on three downstream tasks: named entity recognition (NER), sentiment analysis (SA), and question answering (QA). All model parameters are updated during fine-tuning, with a task-specific classification head on top of the encoder. Evaluation covers NER and QA in Amharic and Tigrinya and SA in Amharic.

[†]XLM-R supports inputs of up to 512 tokens; a 256-token limit is used for computational efficiency. For QA, longer contexts are split into overlapping windows (stride 128), and the possible effect of the limit on QA performance is discussed in Section 5.4.

For NER and SA, we report accuracy as the primary metric.[2] For QA, we report both Exact Match (EM) and F1 scores.

# 4 Experimental Setup

## 4.1 Model Architecture

VEXMLM builds on XLM-RoBERTa-base's encoder architecture, retaining 12 Transformer layers, hidden size of 768, 12 attention heads, GELU activation, and layer normalization ($\epsilon = 1 \times 10^{-5}$). The key change lies in targeted vocabulary expansion: we expand the vocabulary from 250,002 to 280,002 entries by adding 30,000 Ge'ez-script tokens, resulting in 301,365,186 parameters (model card); relative to XLM-R, the expansion adds 30,000 embedding rows of dimension 768. These new vocabulary entries are initialized using mean-based embedding initialization. For reference, Glot500 has 395M parameters and a 400K vocabulary (ImaniGooghari et al., 2023).

## 4.2 Training Configuration

Table 1 lists all settings. Continued pretraining runs for 60 epochs with a learning rate of $5 \times 10^{-5}$, decayed linearly after a 6% warmup; we release the checkpoint with the lowest validation loss, reached at epoch 56 (step 24,808; validation loss 3.7120). Fine-tuning uses a learning rate of $2 \times 10^{-5}$ with a 10% warmup followed by linear decay over 4 epochs. Both stages use sequences of maximum length 256, batch size 32, AdamW optimizer ($\beta_1$=0.9, $\beta_2$=0.999, $\epsilon$=$1 \times 10^{-8}$) with weight decay 0.01, gradient clipping at 1.0, and bf16 mixed-precision training on a single NVIDIA A100 80GB PCIe (pretraining). During both stages, all parameters are trainable; pretraining uses 0.15 MLM probability. For NER and SA, the fine-tuning epoch with the best validation score (macro-F1 for NER, accuracy for SA) is kept; for QA, the final epoch is used. Each VEXMLM downstream configuration is fine-tuned with five seeds (42–46), and we report the mean and standard deviation.

## 4.3 Datasets and Tasks

**Pretraining data.** Continued pretraining uses one Amharic and one Tigrinya monolingual text file drawn from a prepared collection whose provenance is undocumented. The files correspond to no named corpus available to us, and inspection shows a mixture of religious translations and general web prose. Their licence is consequently unknown and we do not redistribute them. After NFC normalization, removal of lines shorter than 10 characters or below 50% Ge'ez characters, and exact deduplication, the Amharic file (200,002 lines) yields 129,402 lines and the Tigrinya file (200,000 lines) 140,583 lines; 2% of each is held out as a development set (2,588 and 2,811 sentences), which we use for the intrinsic tokenizer evaluation, together with the words of the development set that do not occur in the training lines (2,414 Amharic and 3,021 Tigrinya OOV words). Stage 1 uses the remaining training lines, from which a further 1% (2,645 lines) is held out for validation.

We evaluate VEXMLM on NER and QA in Amharic and Tigrinya and on SA in Amharic. TIGQA, which lacks a standard split, is divided 80:10:10 by context (not by question, so no passage appears in two splits) with seed 42, giving 644/67/86 questions over 292/36/37 contexts. We use the 797 span-extractable question–answer pairs over 365 contexts: 1,039 abstractive and 272 unanswerable items are excluded, as an `answer_start` of $-1$ indicates that the answer does not occur verbatim in the context and so cannot be scored by span-extraction EM/F1. These counts sum to 2,108 of the 2,685 pairs in the published release (Teklehaymanot et al., 2024); the provenance of the remaining items is not documented in our copy of the data (`datasets/cards/tigqa.md`). Our TIGQA scores are therefore obtained on this span-extractable subset and are not directly comparable with results on the full release.

[2]Accuracy is used for NER to maintain consistency with SA; however, we acknowledge that macro-F1 is the standard metric for NER evaluation and report macro-F1 and entity-level F1 in Table 6 in the Appendix.

**Sentiment Analysis (SA).** We use the Amharic portion of AfriSenti (Muhammad et al., 2023), a multilingual Twitter corpus with three-way sentiment labels (positive, negative, neutral) that covers 14 African languages (5,984/1,497/1,999 train/dev/test tweets). This benchmark tests VEXMLM's ability to capture sentiment signals in noisy social media text.

**Named Entity Recognition (NER).** We evaluate on the Amharic portion of MasakhaNER (Adelani et al., 2021) (1,750/250/500 sentences), with BIO-tagged PER, LOC, ORG and DATE entities, and on the Tigrinya NER dataset (Yohannes and Amagasa, 2022) (4,562/570/571 sentences), which additionally tags MISC entities. NER and SA scores are computed on the test splits.

**Question Answering (QA).** We use TIGQA (Teklehaymanot et al., 2024), an expert-annotated extractive QA dataset in Tigrinya (644/67/86 questions), and AmQA (Taffa et al., 2024), an extractive QA corpus for Amharic (1,723/600/299 questions in the official release). QA scores are computed on the development splits, which our fine-tuning script uses for QA evaluation; because QA keeps the final fine-tuning epoch, these splits play no role in model selection and serve as held-out data. In the official AmQA release the development split is larger than the test split.

### 4.4 Baselines

We compare VEXMLM against XLM-R (Conneau et al., 2020), the base model from which VEXMLM is derived, fine-tuned with identical hyperparameters and evaluation protocols. This baseline isolates the contribution of vocabulary expansion and continued pretraining. The XLM-R baseline of Table 4 and Table 6 is a single run (seed 42); the XLM-R baseline of the OOV evaluation (Table 3) and the ablation (Table 5) uses five seeds (42–46), like the other ablation arms. For the intrinsic tokenizer evaluation we additionally compare against the tokenizer of Glot500 (ImaniGooghari et al., 2023), which provides a comparison against scale-oriented multilingual pretraining without Ge'ez-script-specific optimization.

Additional African NLP models (e.g., AfroXLMR, AfriBERTa) are excluded from the comparison; their positioning relative to this work is discussed in Section 5.4.

### 4.5 Ablation Study Design

We conduct ablation experiments on Tigrinya NER to isolate the contribution of each component. Tigrinya is selected as the ablation target given its representation in both the expansion vocabulary and downstream tasks. We compare four models, each fine-tuned on Tigrinya NER with the Stage 2 settings and five seeds:

1. **XLM-R (baseline)**: No vocabulary expansion, no continued pretraining.

2. **+VocabExp (Random Init)**: Vocabulary expanded with 30,000 new tokens, initialized from a zero-mean normal distribution with the standard deviation of the pretrained embeddings (0.21). No continued pretraining.

3. **+VocabExp (Mean Init)**: Vocabulary expanded with 30,000 new tokens, mean-initialized (Equation 2). No continued pretraining.

4. **VEXMLM (Full)**: Mean-initialized vocabulary expansion + continued pretraining.

Models 2–4 share the same expanded tokenizer. The metric is token-level accuracy on out-of-vocabulary (OOV) words of the Tigrinya NER test split, with the repository's definition (`evaluation/run_ner_oov.py`): a word is OOV when the XLM-R tokenizer maps it to [UNK], cannot reproduce it on decoding, or splits it into more pieces than the expanded tokenizer. The set (3,491 of the 4,677 word types) is thus defined relative to the expanded tokenizer and includes words that XLM-R can represent but fragments more than VEXMLM; it is identical for all four models.

| Metric | Language | XLM-R | Glot500 | VEXMLM |
|---|---|---|---|---|
| Fertility ↓ | Amharic | 2.0692 | 2.0255 | **1.4888** |
| | Tigrinya | 3.1300 | 2.4703 | **1.6928** |
| Compression ↑ | Amharic | 2.2950 | 2.3445 | **3.1896** |
| | Tigrinya | 1.4591 | 1.8487 | **2.6979** |
| Round-trip ↑ | Amharic | 1.0000 | 1.0000 | 1.0000 |
| | Tigrinya | 0.9954 | 1.0000 | 0.9987 |

Table 2: Tokenizer metrics on held-out Amharic and Tigrinya text: fertility (tokens per word, lower is better), compression (characters per token, higher is better) and the share of OOV words (development-set words absent from the training lines) that the tokenizer reproduces exactly on decoding. Bold indicates the best fertility and compression. All three tokenizers are measured on the same development sets; the XLM-R and VEXMLM values are recorded in `results/spmerge_tokenizer_metrics.json` and the Glot500 values in `results/provenance/tokenizer_metrics_2026-08-15.json`, which agree on XLM-R.

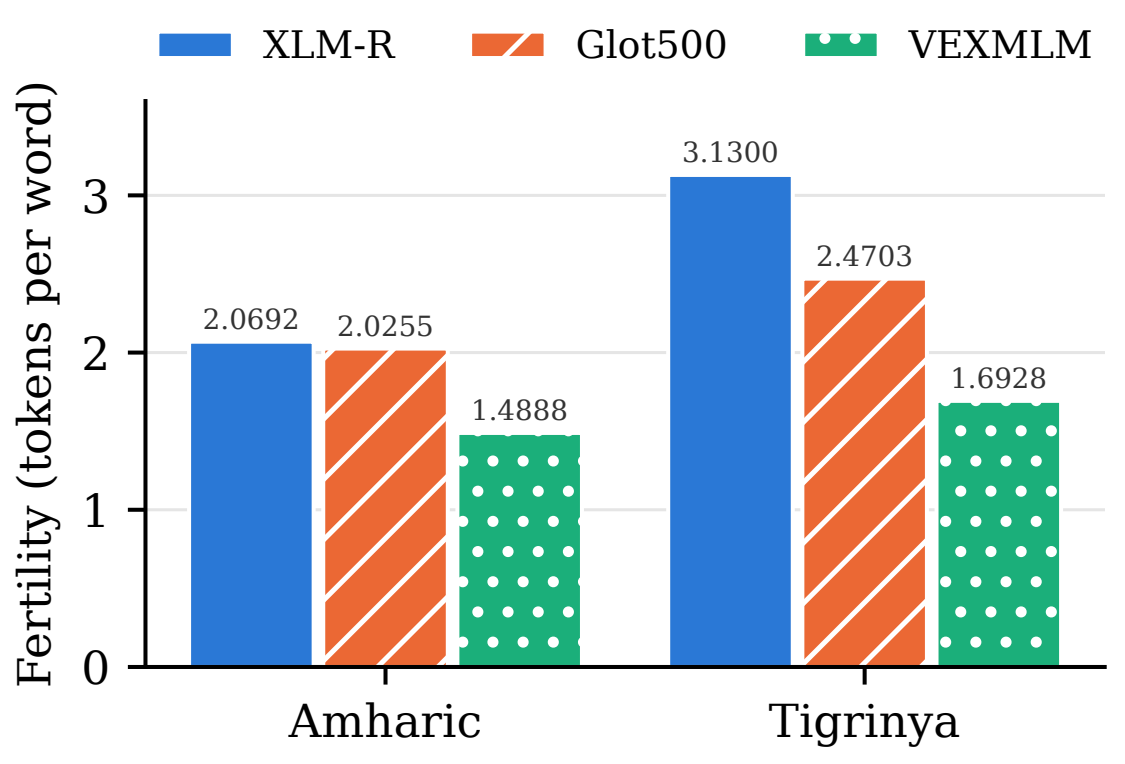

Figure 1: Fertility scores (tokens per word) for Amharic and Tigrinya. VEXMLM reduces fertility by 45.9% for Tigrinya and 28.0% for Amharic relative to XLM-R.

| Language | XLM-R | | VEXMLM | |
|---|---|---|---|---|
| | Non-OOV | OOV | Non-OOV | OOV |
| Tigrinya | 93.27 ± 0.16 | 94.57 ± 0.16 | 94.12 ± 0.14 | **95.66 ± 0.09** |

Table 3: Accuracy (%) in Tigrinya NER (test split) on OOV words (Section 4.5) and on all other words (Non-OOV); mean ± standard deviation over five seeds.

## 5 Results and Evaluation

We evaluate VEXMLM using intrinsic tokenization quality metrics and extrinsic downstream task performance on Amharic and Tigrinya. NER and SA accuracy and F1 are given as fractions to four decimals, QA Exact Match and F1 and the ablation accuracies as percentages to two decimals, and tokenizer metrics to four decimals. VEXMLM scores are means ± sample standard deviations over five seeds, and differences are computed from unrounded means, so they may differ by one unit in the last decimal from differences of the displayed values.

### 5.1 Intrinsic Evaluation: Tokenization Quality

Table 2 compares the tokenizers of XLM-R, Glot500 and VEXMLM on the held-out Amharic and Tigrinya development sentences (2,588 and 2,811). We do not report parity (Petrov et al., 2023), which requires a sentence-aligned parallel corpus that was not available.

**Fertility.** Fertility (Rust et al., 2021) measures the average number of tokens produced per input word; lower values indicate more efficient tokenization. Figure 1 shows that VEXMLM achieves substantially lower fertility than XLM-R for Tigrinya (3.1300 → 1.6928, a 45.9% reduction) and Amharic (2.0692 → 1.4888, a 28.0% reduction). Glot500 achieves intermediate fertility on both Ge'ez-script languages (2.0255 and 2.4703). This reduction confirms that vocabulary expansion and integration effectively address the tokenization inefficiency observed in XLM-R for Ge'ez-script languages.

**Compression.** Compression (Goldman et al., 2024) measures characters per token; higher values indicate more efficient encoding. VEXMLM achieves higher compression for Amharic and Tigrinya than XLM-R and Glot500, in line with its fertility improvements. All three tokenizers reproduce nearly all OOV words exactly on decoding (Table 2).

**Vocabulary coverage.** VEXMLM's vocabulary of 280,002 entries preserves all original XLM-R entries, so coverage for the 100 languages supported by the base model is unchanged.

**OOV words.** Table 3 reports Tigrinya NER accuracy on OOV words (Section 4.5) and on all other words. VEXMLM improves accuracy by +1.09 points on OOV words and by +0.85 points on the other words. For both models, accuracy on OOV words is higher than on the other words.

### 5.2 Extrinsic Evaluation: Downstream Task Performance

Table 4 reports performance on all three downstream tasks. On NER, VEXMLM achieves higher accuracy than XLM-R on both MasakhaNER Amharic (+0.0046) and Tigrinya NER (+0.0084), and higher macro-F1 on both (+0.0314 and +0.0344;

| Task (Metric) | Dataset | XLM-R* | VEXMLM |
|---|---|---|---|
| SA (Accuracy) | AfriSenti (Amharic) | 0.5073 | 0.4978 ± 0.0331 |
| NER (Accuracy) | MasakhaNER (Amharic) | 0.9366 | **0.9413 ± 0.0026** |
| | Tigrinya NER (Tigrinya) | 0.9431 | **0.9515 ± 0.0005** |
| QA (EM, %) | AmQA (Amharic) | **41.50** | 32.57 ± 0.77 |
| | TIGQA (Tigrinya) | **13.43** | 2.39 ± 0.82 |
| QA (F1, %) | AmQA (Amharic) | **58.46** | 48.85 ± 0.96 |
| | TIGQA (Tigrinya) | **27.90** | 9.76 ± 0.97 |

Table 4: Downstream task performance. SA and NER report accuracy (fraction) on the test splits; QA reports Exact Match (EM) and F1 (%) on the development splits. VEXMLM: mean ± standard deviation over five seeds. *XLM-R: single run (seed 42). Bold marks the higher score where the difference exceeds one standard deviation of VEXMLM.

Table 6), larger margins than token accuracy shows; entity-level F1 is higher on Tigrinya (+0.0464) and comparable on Amharic, where the difference (+0.0099) is within one standard deviation. On QA, VEXMLM scores below XLM-R on both datasets: by 8.93 EM and 9.61 F1 points on AmQA, and by 11.04 EM and 18.13 F1 points on TIGQA. On SA, the difference in accuracy (-0.0094) is within one standard deviation of VEXMLM's scores, so the two models are comparable. Because the XLM-R baseline is a single run, its own seed variance is unknown, and small differences should be read with caution.

### 5.3 Ablation Study

Table 5 reports ablation results for Tigrinya NER OOV accuracy, isolating the contribution of each component. Vocabulary expansion with random initialization lowers this accuracy by 7.52 points relative to the baseline; without continued pretraining, the new embedding rows receive no training before fine-tuning. Mean initialization recovers 0.59 points relative to random initialization. The full VEXMLM configuration (mean initialization + continued pretraining) achieves the best accuracy, with continued pretraining providing the largest single gain (8.02 points) and the full model finishing +1.09 points above the baseline. Updating the model's representations through MLM on Ge'ez-script corpora is therefore the most critical component of the pipeline. The configurations without continued pretraining also receive less training on Amharic and Tigrinya text, so part of this gain may reflect the additional training rather than the adaptation of the new embeddings; separating the two would require an unexpanded model given the same continued pretraining, which we did not run.

| Configuration | OOV Acc. (%) | $\Delta$ |
|---|---|---|
| XLM-R baseline | 94.57 ± 0.16 | – |
| + VocabExp (Random Init) | 87.04 ± 0.20 | -7.52 |
| + VocabExp (Mean Init) | 87.63 ± 0.14 | +0.59 |
| + Continued Pretraining (VEXMLM) | **95.66 ± 0.09** | +8.02 |

Table 5: Ablation study: Tigrinya NER OOV accuracy under different configurations (mean ± standard deviation over five seeds). Rows 2 and 3 expand the vocabulary with random or mean initialization; row 4 adds continued pretraining to row 3; $\Delta$ is the change from the previous row, computed from unrounded means. Values as in the repository's `results/table5_ablation.csv`.

### 5.4 Discussion

The results show that VEXMLM's clearest improvement is in tokenization efficiency: targeted Ge'ez-script subword augmentation reduces fertility relative to XLM-R by 45.9% for Tigrinya and 28.0% for Amharic, and yields lower fertility than Glot500 on both languages. The downstream effect of this efficiency depends on the task. On NER, VEXMLM improves over XLM-R, and the gain on Tigrinya holds for both OOV and other words. On extractive QA, VEXMLM scores below XLM-R on both datasets. The cause of this gap is not analysed in this work. On SA, the two models are comparable.

The ablation shows that vocabulary expansion by itself is harmful: new embeddings, whether random or mean-initialized, lower OOV accuracy until continued MLM pretraining adapts them. Mean initialization is a better starting point than random initialization, but the difference between them is small compared with the effect of continued pretraining.

Comparing with AfroXLMR and AfriBERTa is complicated by differing language coverage and evaluation protocols, and a direct, controlled comparison constitutes important future work. The 256-token sequence limit, under which long QA contexts are read in overlapping windows, may also constrain QA performance on long documents; future work should investigate full-length training.

## 6 Conclusion

We present VEXMLM, a vocabulary-extended variant of XLM-R designed to address excessive subword fragmentation in Ge'ez-script languages through: (i) language-specific SentencePiece tokenization for Amharic and Tigrinya; (ii) augmentation with 30,000 Ge'ez-derived subword tokens; (iii) mean-based embedding initialization for compatibility with the pretrained embedding space; and

(iv) a two-stage adaptation framework in which the vocabulary-extended model is first adapted through continued MLM pretraining on Amharic and Tigrinya corpora and then fine-tuned for downstream tasks.

VEXMLM reduces tokenizer fertility relative to XLM-R by 28.0% for Amharic and 45.9% for Tigrinya, below that of Glot500. Downstream, it improves named entity recognition over XLM-R, is comparable on sentiment analysis, and scores below XLM-R on extractive question answering. The ablation indicates that continued MLM pretraining is the primary driver of the gains: vocabulary expansion alone lowers accuracy on out-of-vocabulary words, and mean-based initialization is only a slightly better starting point than random initialization.

Overall, our findings show that language-aware vocabulary expansion makes the tokenization of underrepresented Ge'ez-script languages substantially more efficient, but that its downstream benefit is task-dependent and requires adapting the new embeddings through continued pretraining.

## Ethical Considerations

This work aims to improve NLP capabilities for historically underserved languages in African communities. The downstream benchmarks we use are publicly available. The origin and licence of the monolingual pretraining text are not documented (Section 4.3); we therefore do not redistribute it. We do not introduce new data collection; potential biases in the source corpora may propagate to VEXMLM's representations.

## 7 Limitations and Future Directions

Several limitations remain. First, the vocabulary expansion and tokenization strategy were developed specifically for Amharic and Tigrinya, and the approach has not been evaluated on other Ge'ez-script languages such as Tigre, Harari, or Classical Ge'ez, nor on other African languages. Second, the XLM-R baseline of the downstream task tables (Table 4, Table 6) was run with a single seed, so its seed variance there is unknown; the VEXMLM results, and both models in the OOV and ablation tables, are averaged over five seeds. Third, the OOV word set used in Table 3 and Table 5 is defined relative to the expanded tokenizer and includes words that XLM-R can represent but fragments more than VEXMLM, so it does not measure unseen words only. The ablation also does not separate the adaptation of the new embeddings from the additional training that continued pretraining provides, because no unexpanded model received the same continued pretraining. Fourth, the pretraining text is of undocumented origin, which limits the data statement we can give and prevents redistribution. Fifth, the proposed tokenizer remains primarily frequency-driven and does not explicitly model the rich morphological structure of Ge'ez-script languages. Finally, the QA development sets are small (67 TIGQA questions).

Our findings also suggest that effective vocabulary integration is as important as vocabulary expansion itself: the expanded vocabulary helps only after continued pretraining.

Future work will investigate vocabulary augmentation for additional Ge'ez-script languages, morphology-aware tokenization strategies, hybrid character-byte representations, a multi-seed XLM-R baseline, and the causes of the lower QA scores. Future studies should also examine alternative vocabulary integration methods and controlled comparisons with African-focused multilingual models such as AfroXLMR and AfriBERTa under a unified evaluation framework.

## 8 Additional Downstream Metrics

Table 6 reports macro-F1 for NER and SA and entity-level F1 for NER, complementing the accuracy scores in Table 4.

| **Task (Metric)** | **Dataset** | **XLM-R*** | **VEXMLM** |
|---|---|---|---|
| SA (Macro-F1) | AfriSenti (Amharic) | 0.5036 | 0.4971 ± 0.0193 |
| NER (Macro-F1) | MasakhaNER (Amharic) | 0.7109 | **0.7423 ± 0.0122** |
| | Tigrinya NER (Tigrinya) | 0.7875 | **0.8219 ± 0.0069** |
| NER (Entity-F1) | MasakhaNER (Amharic) | 0.6248 | 0.6347 ± 0.0148 |
| | Tigrinya NER (Tigrinya) | 0.6817 | **0.7282 ± 0.0079** |

Table 6: Macro-F1 and entity-level F1 (fractions) on the test splits. VEXMLM: mean ± standard deviation over five seeds. *XLM-R: single run (seed 42). Bold marks the higher score where the difference exceeds one standard deviation of VEXMLM.

## 9 Reproducibility

Code, configurations and result files are available at `https://github.com/hailaykidu/VEXMLM` (release tag `v1.1-correction`); the released model is `https://huggingface.co/Hailay/VEXMLM`. Every result number, table and figure in this paper is generated from the run outputs by `scripts/make_paper_artifacts.py`, which reads the aggregated metrics produced by `evaluation/export_spm_results.py` (`results/downstream_task_metrics.csv`) and `evaluation/aggregate_ablation.py` (`results/table5_ablation.csv`). `REPRODUCE.md` maps each table and figure to the script, configuration and command that produced it.